\pdfoutput=1
\documentclass[11pt]{article}
\usepackage[preprint]{acl}
\usepackage{times}
\usepackage{latexsym}
\usepackage[T1]{fontenc}
\usepackage[utf8]{inputenc}
\usepackage{microtype}
\usepackage{inconsolata}
\usepackage{graphicx}
\usepackage{amsmath}%
\usepackage{amssymb}%
\usepackage{pifont}%
\usepackage{arydshln}
\usepackage{url}

\usepackage{tcolorbox}
\tcbuselibrary{listingsutf8}
\usepackage{listings}
\usepackage{courier}
\usepackage{caption}
\usepackage{bm}
\newcommand{\cmark}{\ding{51}}%
\newcommand{\xmark}{\ding{55}}%
\newcommand{\debertabase}{${\textnormal{DeBERTa}}_{\textnormal{BASE}}$ }
\newcommand{\debertabasenospace}{${\textnormal{DeBERTa}}_{\textnormal{BASE}}$}
\newcommand{\debertalarge}{${\textnormal{DeBERTa}}_{\textnormal{LARGE}}$ }
\newcommand{\debertalargenospace}{${\textnormal{DeBERTa}}_{\textnormal{LARGE}}$}
\newcommand{\debertalargecompact}{${\textnormal{DeBERTa}}_{\textnormal{LRG}}$ }

\title{Extractive Fact Decomposition for Interpretable Natural Language Inference in one Forward Pass}

\author{Nicholas Popovič \and Michael Färber\\
TU Dresden \& ScaDS.AI Dresden/Leipzig, Germany\\
\small\texttt{\{nicholas.popovic,michael.faerber\}@tu-dresden.de}}

\begin{document}
\maketitle
\begin{abstract}

Recent works in Natural Language Inference (NLI) and related tasks, such as automated fact-checking, employ atomic fact decomposition to enhance interpretability and robustness. For this, existing methods rely on resource-intensive generative large language models (LLMs) to perform decomposition. We propose JEDI, an encoder-only architecture that jointly performs extractive atomic fact decomposition and interpretable inference without requiring generative models during inference. To facilitate training, we produce a large corpus of synthetic rationales covering multiple NLI benchmarks. Experimental results demonstrate that JEDI achieves competitive accuracy in distribution and significantly improves robustness out of distribution and in adversarial settings over models based solely on extractive rationale supervision. Our findings show that interpretability and robust generalization in NLI can be realized using encoder-only architectures and synthetic rationales.\footnote{Code and data: \url{https://jedi.nicpopovic.com}}

\end{abstract}

\begin{figure}[!ht]
    \centering
    \includegraphics[width=1\linewidth]{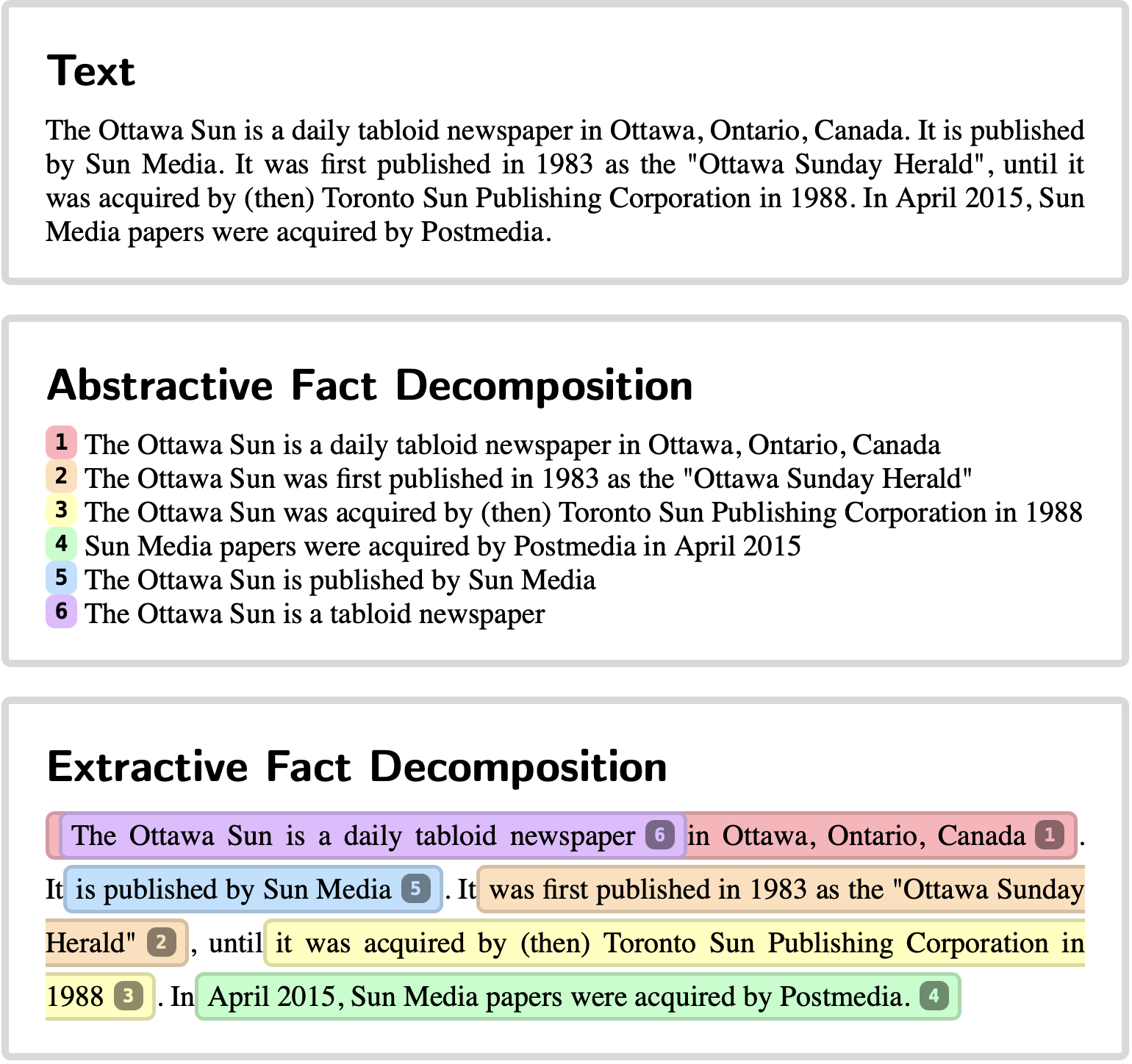}
    \caption{Illustration of abstractive and extractive fact decomposition for an example premise from the ANLI dataset. Abstractive atomic facts are based on the generated facts provided by \citet{stacey-etal-2024-atomic}. Extractive facts are shown as colored spans corresponding to the contents of the abstractive atomic facts.}
    \label{fig:atomic-facts}
\end{figure}

\section{Introduction}

Natural language inference (NLI) tasks \cite{giampiccolo-etal-2007-third, bowman-etal-2015-large} require models to determine whether a given hypothesis logically follows from a premise. While state-of-the-art NLI models have achieved high accuracy, their decision-making processes often remain opaque. This has motivated a growing body of research focused on building interpretable NLI systems that not only predict a label but also justify their predictions in a transparent and faithful way \cite{NIPS2018_8163,deyoung-etal-2020-eraser,stacey-etal-2022-logical,stacey-etal-2024-atomic}.\\

A common approach to interpretability in NLI involves extractive rationales, where the model highlights parts of the premise or hypothesis that support its decision. While easily interpretable, such rationales often fail to capture the underlying logical structure of inference and can obscure shallow pattern matching \cite{mccoy-etal-2019-right}. In response to these limitations, recent work has turned to atomic fact decomposition, which breaks the premise into minimal, semantically coherent sub-facts (atoms) against which the hypothesis is then validated individually \cite{stacey-etal-2024-atomic}. By adding symbolic reasoning over these atomic units, this approach improves transparency, robustness and more closely reflects the compositional reasoning involved in NLI.\\

However, atomic fact decomposition currently relies on generating atomic facts using large language models (LLMs), which introduces additional computational overhead. Further, LLMs have been shown to hallucinate inaccuracies that can negatively impact inference and may be difficult to identify since atomic facts are not immediately traceable to explicit premise spans. Such hallucinations are especially problematic in longer documents, making verification challenging and hindering scalability. This raises the central question of our work:
\textit{Can atomic fact decomposition be distilled into encoder-only architectures to enable fast, scalable, and faithfully interpretable NLI without requiring LLMs at inference time?}
To answer this, we propose reframing atomic fact decomposition as an extractive task, similar to existing extractive rationale-based approaches. Rather than generating textual statements, our model identifies spans in the premise corresponding directly to atomic facts. These spans are then classified with respect to the hypothesis using logical rules, enabling fact-level interpretability in a single encoder forward pass.\\

Our contributions in this paper are as follows:
\begin{itemize}
    \item We propose an encoder-only architecture (JEDI\footnote{JEDI: Joint Encoder for Decomposition and Inference}) that performs extractive atomic fact decomposition and logical inference jointly, enabling interpretable NLI without text generation during inference.
    \item We construct a large corpus of synthetic rationales (SYRP\footnote{SYRP: SYnthetic Rationales for Premises}) to supervise span-level extraction in the absence of annotated data. In addition to the annotations required for this paper, we provide synthetic rationales across seven additional datasets to facilitate future research in interpretable NLI.
    \item We demonstrate that our approach produces competitive results both in distribution and out of distribution while offering fine-grained, faithfully interpretable predictions grounded explicitly in the premise.
\end{itemize}

\section{Related Work}

\subsection{Extractive Rationales in NLI}
Natural Language Inference (NLI) \cite{giampiccolo-etal-2007-third, bowman-etal-2015-large} has long served as a key task for evaluating model reasoning capabilities. Early work on interpretability emphasized extractive rationale methods, which highlight input spans presumed to justify model predictions \cite{deyoung-etal-2020-eraser}. For instance, e-SNLI \cite{NIPS2018_8163} introduced human-annotated rationales aligned with entailment labels to support more transparent decision-making.
However, subsequent studies, exemplified by the work of \citet{chen-etal-2022-rationalization}, revealed that rationale supervision leaves models vulnerable to adversarial attacks, such as superficial pattern matching \cite{mccoy-etal-2019-right}.
This motivates a shift toward more structured and granular reasoning frameworks.\\

\subsection{Atomic Fact Decomposition}
Recent approaches decompose NLI examples into atomic units to support finer-grained inference. 
While \citet{stacey-etal-2022-logical} focused on span-level predictions grounded in noun phrase segmentation of the hypothesis, their subsequent work \cite{stacey-etal-2024-atomic} proposes training models using atomic facts derived from premise decomposition via LLMs. These methods aim to disentangle model reasoning from shallow heuristics by isolating semantically meaningful units and applying rule-based reasoning.
Decomposition-based strategies have also been explored in related domains such as summarization \cite{yang-etal-2024-fizz}, fact-checking \cite{min-etal-2023-factscore}, and claim verification \cite{kamoi-etal-2023-wice,chen-etal-2024-complex}. The reliance on generative LLMs for decomposition introduces computational overhead, limited scalability, and potential inaccuracies (hallucinations) in inference, motivating our exploration of alternative extraction-based approaches.\\

\subsection{Joint Architectures}
Recently, \citet{lu2025optimizingdecompositionoptimalclaim} argue that pipeline architectures involving fact decomposition suffer from error propagation and advocate for joint training to reduce reliance on intermediate supervision.
In this work, we draw a parallel to a different natural language processing task in which joint architectures are common: We take inspiration from joint entity and relation extraction \cite{eberts-ulges-2021-end,zhou2021atlop,popovic-etal-2022-aifb,hennen-etal-2024-iter} where span extraction and complex classification are at the core of the task. We propose an extractive approach to fact decomposition, enabling a single encoder-based model to jointly learn both decomposition and inference. Joint models, however, require rich annotated data to effectively learn both decomposition and inference. Given the lack of existing annotations at this granularity, we create a synthetic dataset spanning multiple NLI benchmarks, generated via LLMs, to facilitate training and evaluation.

\section{Extractive Fact Decomposition}
\label{sec:extractive-fact-decomposition}
The core premise of this work is to view the process of atomic fact decomposition from an extractive point of view, rather than the abstractive approach seen in recent works \cite{stacey-etal-2024-atomic, yang-etal-2024-fizz, min-etal-2023-factscore, chen-etal-2024-complex}, which make use of generative models.
Figure \ref{fig:atomic-facts} provides an illustration of the two annotation types.
While abstractive atomic fact decomposition has demonstrated improved robustness compared to extractive rationale-based interpretability methods, this work explores the hypothesis that the observed robustness improvements may not inherently depend on abstraction via generation, but rather on the structured reasoning over clearly defined semantic units, and therefore also be achievable via extractive methods.\\

Further motivation for exploring an extractive framing stems from several practical and methodological considerations: Extractive rationales, as demonstrated in prior interpretability research \cite{deyoung-etal-2020-eraser, NIPS2018_8163}, provide explicit pointers to the relevant portions of the input text (the premise, for NLI), which is especially valuable when dealing with long or complex contexts.
By providing a more transparent and readily verifiable means to trace predictions directly to explicit spans in the input, an extractive approach can help reduce risks introduced by potential hallucinations associated with generated facts.
Finally, extractive rationales lend themselves more naturally to encoder-only architectures. These models are typically significantly more lightweight than generative ones and continue to be used in many downstream natural language understanding (NLU) tasks for their efficiency.\\

\begin{figure}[!ht]
    \centering
    \includegraphics[width=1\linewidth]{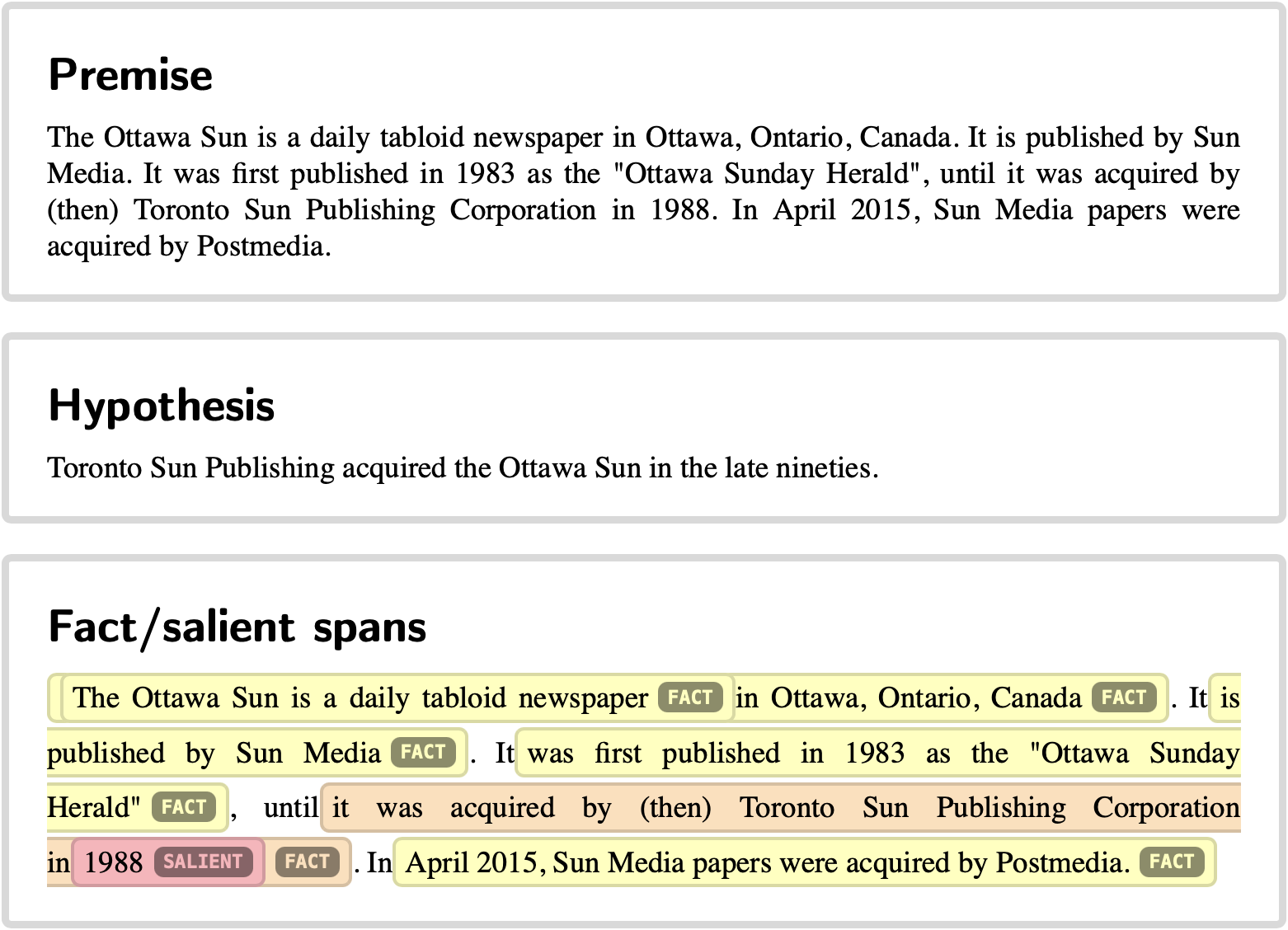}
    \caption{Illustration of fact spans (yellow/orange) and salient spans (red) for an example premise and hypothesis from the ANLI dataset. The example contains a single salient span highlighting information contradicting the hypothesis. During training, we treat fact spans which contain salient spans (shown in orange) as salient spans as well.}
    \label{fig:span-types}
\end{figure}

\section{SYRP: SYnthetic Rationales for Premises}
\label{sec:syrp}
Data in existing NLI datasets typically consists of premise, hypothesis, and label.
For NLI via extractive fact decomposition we require two additional types of span annotations shown in Figure \ref{fig:span-types}, which we refer to as \textit{fact spans} and \textit{salient spans}.
Fact spans define spans in the premise corresponding to individual atomic facts, while salient spans correspond to the spans in the premise which are most relevant to the predicted label (equivalent to extractive rationales).\\

\begin{figure*}[!ht]
    \centering
    \includegraphics[width=1\linewidth]{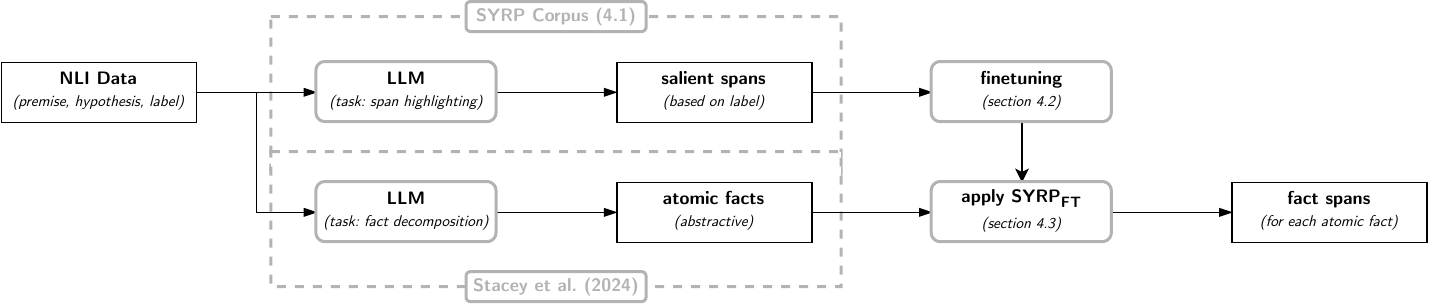}
    \caption{Overview of the pipeline for data collection as described in Section 4.}
    \label{fig:datapipeline}
\end{figure*}

To address this annotation gap, we synthetically create fact and salient span annotations in a multi-step process shown in Figure \ref{fig:datapipeline}:
First, we create synthetic salient spans for several NLI benchmarks (Section \ref{sec:syrp-corpus}).
Using the atomic facts \citet{stacey-etal-2024-atomic} generated for ANLI \cite{nie-etal-2020-adversarial}, a challenging benchmark central to recent research in NLI, and a token classification model trained on our synthetic rationales (Section \ref{sec:syrp-ft}), we then create annotations of fact spans (Section \ref{sec:bootstrap}).\\

\subsection{SYRP corpus}
\label{sec:syrp-corpus}
We create synthetic salient span annotations, for the training and development splits of the ANLI dataset \cite{nie-etal-2020-adversarial} using an LLM\footnote{We selected Qwen2.5-32B-Instruct-GPTQ-Int4 based on performance and efficiency.} and carefully designed prompt templates detailed in Appendix \ref{sec:appendix-syrp}, Figures \ref{fig:llm-prompt-entailed} and \ref{fig:llm-prompt-contradicted}.
The final configuration of model and prompt templates were chosen based on an evaluation of 15 different models and 20 different prompt templates, conducted on manually annotated data.
More details can be found in Appendix \ref{sec:appendix-syrp}.\\

Broadly, we frame the annotation task by providing an instruction tuned model with the premise, hypothesis, and gold label.
This way the model no longer has to perform the full task of verifying the hypothesis, but only needs to provide relevant spans.
This means that even a model which does not achieve state-of-the-art performance on NLI can produce rationales for a given data sample.
We evaluate annotation quality using intersection-over-union (IoU) with manually annotated spans, achieving an IoU of 69\%, indicating substantial agreement (IoU > 50\% is considered indicative of good agreement by \citet{deyoung-etal-2020-eraser}).
To ensure annotations reflected genuine task comprehension, we additionally evaluated accompanying natural language explanations. We found only 2 out of 30 explanations to be of low quality, further validating robustness.\footnote{We do not use the natural language explanations in the remainder of this work, but evaluated them in the initial model selection as a proxy for task comprehension and include the generations as part of the SYRP corpus for future research.}\\

While our primary focus is ANLI, we have additionally generated a corpus comprising roughly one million annotated samples across eight NLI benchmarks, publicly available to support future research in interpretable NLI.
Statistics for this dataset, the SYRP corpus, can be found in Appendix \ref{sec:appendix-syrp-stats}.

\subsection{Token Classification Models (SYRP$_\text{FT}$)}
\label{sec:syrp-ft}
Using the synthetic rationales produced for ANLI above, we finetune encoder-based token-classification models, SYRP$_\text{FT}$.
For this, we pass premise and hypothesis to the encoder with a leading CLS token and a separator token (SEP) to mark the end of the premise.
Each token in the premise is classified as either neutral (non-salient), entailed (supporting entailment), or contradicted (supporting contradiction). During inference, we follow logic from prior work \cite{stacey-etal-2024-atomic}, predicting contradiction if any contradicted tokens exist, entailment if entailed tokens exist without contradictions, and neutral otherwise.
As a result, any predictions produced by SYRP$_\text{FT}$ are traceable to salient tokens in the premise.

\subsection{Span-Level Supervision from Generated Atomic Facts}
\label{sec:bootstrap}
With salient span annotations in place (i.e., rationales decisive to entailments and contradictions), the remaining supervision signal we require is for fact spans (spans representing atomic facts, including those that are not directly relevant to the hypothesis), as illustrated in Figures \ref{fig:atomic-facts} and \ref{fig:span-types}.\\

To obtain these, we convert the atomic facts generated by \citet{stacey-etal-2024-atomic} for ANLI into fact spans. We use SYRP$_\text{FT}$\footnote{We used \debertalarge as the base encoder for this step.} to identify the salient tokens in the premise that supports each generated fact (supplied to the model as a hypothesis) and convert these to coherent spans. In cases where no fact span corresponding to an atomic fact was detected, we discarded the generated fact as this may indicate a hallucinated statement.\\

This results in a dataset that includes not only salient spans for interpretable NLI, but also fact spans corresponding to individual atomic facts in the premise.
\begin{figure*}[!ht]
    \centering
    \includegraphics[width=1\linewidth]{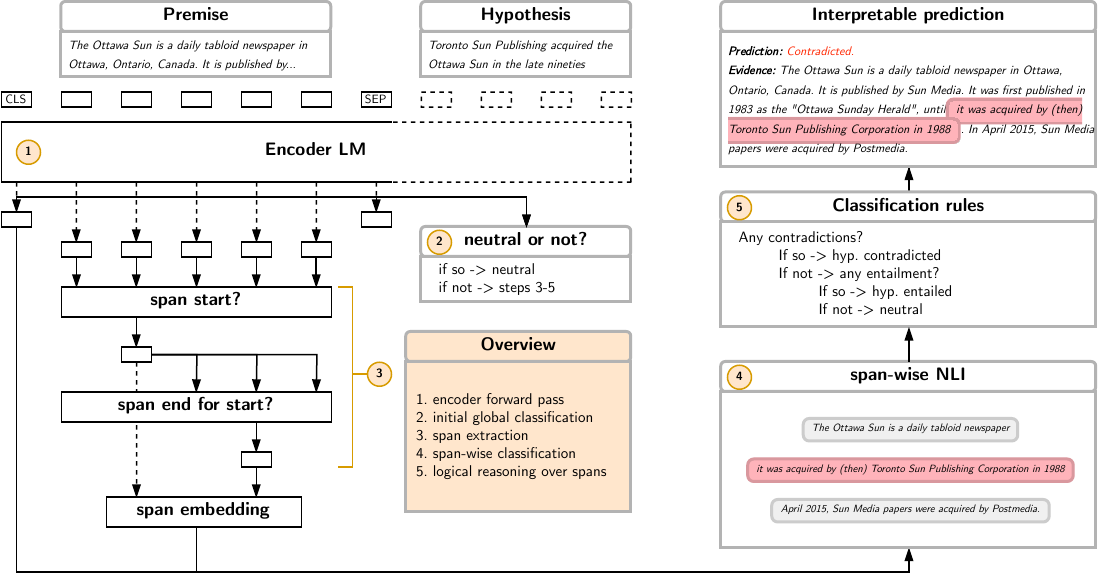}
    \caption{Overview of the proposed architecture (JEDI) for performing fact-level span extraction and logical reasoning to perform interpretable natural language inference in a single forward pass.
    }
    \label{fig:architecture}
\end{figure*}

\section{JEDI: Joint Encoder for Decomposition and Inference}

In this section, we describe our model architecture for joint fact decomposition and interpretable natural language inference, an overview of which is provided in Figure \ref{fig:architecture}.
We combine modeling approaches from information extraction for example, \cite{eberts-ulges-2021-end,zhou2021atlop,hennen-etal-2024-iter}, where efficient and effective span extraction and classification is at the heart of the task, with the logical rule-based framework for atomic fact-based natural language inference proposed by \citet{stacey-etal-2024-atomic}.
This results in an encoder-only model architecture which performs premise decomposition and atom-level interpretable classification while requiring only a single forward pass and no LLM at inference time.
We refer to the architecture by the acronym JEDI for Joint Encoder for Decomposition and Inference.

\subsection{Encoder forward pass}
Given a premise and a hypothesis, we pass both to the encoder with a leading CLS token and a separator token (SEP) to mark the end of the premise.
For the subsequent computations we discard any embeddings produced for hypothesis tokens and use only the CLS token embedding, $e_{\text{CLS}}$, the token-wise premise embeddings $[e_{\text{p,1}},...,e_{\text{p,n}}]$, and the SEP token embedding, $e_{\text{SEP}}$.
\subsection{Initial global classification}
Taking inspiration from relation extraction models \cite{zhou2021atlop}, we perform an initial global classification $P(x|e_{\text{CLS}}, e_{\text{SEP}})$ using a group bilinear layer\footnote{A variant of bilinear classifiers which reduces the number of parameters by splitting the embedding dimensions into $k$ equal-sized groups \cite{zhou2021atlop}.} applied to the embeddings $e_{\text{CLS}}$ and $e_{\text{SEP}}$ as follows:
\begin{align}
    \left[\bm{z}_\text{CLS}^1;...;\bm{z}_\text{CLS}^k \right] &= e_\text{CLS}, \nonumber \\
    \left[\bm{z}_{\text{SEP}}^1;...;\bm{z}_{\text{SEP}}^k \right] &= \tanh \left( e_{\text{SEP}} \right), \nonumber \\
   \mathrm{P}\left(x|e_\text{CLS}, e_\text{SEP}\right) &= \sigma \left( \sum_{m=1}^k \bm{z}_\text{CLS}^{m\intercal} \bm{W}_x^m \bm{z}_{\text{SEP}}^m + b_x \right) \nonumber
\end{align}
where $\bm{W}_r^i\in \mathbb{R}^{d/k \times d/k}$ for $m=1...k$ are model parameters and $P(x|e_{\text{CLS}}, e_{\text{SEP}})$ is the probability that the hypothesis is neutral, entailed, or contradicted.\\

During inference, if this classification is neutral, we end computation here and output neutral as the predicted class.
If, however, the prediction is contradicted or entailed, we proceed with the span extraction.
During training, steps 3-5 are followed even for neutral examples, as we have a supervision signal from the atomic facts.
This is in contrast to purely rationale-based approaches, as no salient spans are available for neutral examples.

\subsection{Span extraction}
In short, for the span extraction, we take inspiration from \citet{liu-etal-2022-autoregressive} and \citet{hennen-etal-2024-iter} by first identifying tokens likely starting an atomic fact, followed by pairing these potential start tokens with possible end tokens, forming candidate atomic fact spans.\\

For each token at index $i$ in the premise we predict the probability that it represents the first token of a span:
\begin{align}
    P(i_\text{is\_left}|e_{\text{p,i}}) &= \sigma \left( e_{\text{p,i}} \bm{W}_\text{is\_left}^\intercal + b_\text{is\_left} \right), \nonumber
\end{align}
and then, for each token at index $j$ in the premise, compute the probability that the span $(i,j)$ represents a relevant span (or \textit{``atomic fact''}):
\begin{align}
    \bm{e}_{i,j} &= [e_{\text{p},i}; e_{\text{p},j}]\bm{W}_\text{red,1}^\intercal + \bm{b}_\text{red,1}, \nonumber \\
    P(i_\text{is\_span}|{e}_{i,j}) &= \sigma\left( \bm{W}_\text{span} \cdot \bm{e}_{i,j} + b_\text{span} \right), \nonumber
\end{align}
where $[e_{\text{p},i}; e_{\text{p},j}] \in \mathbb{R}^{2d}$ is the concatenation of the embeddings of the start and end tokens of the span, $\bm{W}_\text{red,1} \in \mathbb{R}^{h \times 2d}$ and $\bm{b}_\text{red,1} \in \mathbb{R}^{h}$ are parameters of a linear layer used to project the concatenated token embeddings into a fixed-size span representation $\bm{e}_{i,j} \in \mathbb{R}^{h}$, and $\bm{W}_\text{span} \in \mathbb{R}^{1 \times h}$, $b_\text{span} \in \mathbb{R}$ are the parameters of a binary classifier that outputs the probability that the span $(i,j)$ expresses a relevant atomic fact.

\subsection{Span-wise classification}
Next, given the extracted spans which represent fact atoms, we perform span-wise classification to determine whether the hypothesis is neutral, entailed, or contradicted by a given span $(i,j)$. 
After applying a separate reduction using a linear layer:
\begin{align}
    \bm{e}_{i,j} &= [e_{\text{p},i}; e_{\text{p},j}]\bm{W}_\text{red,2}^\intercal + \bm{b}_\text{red,2}, \nonumber
\end{align}
Using the same group bilinear classifier as for the initial global classification, we compute the following:
\begin{align}
    \left[\bm{z}_\text{CLS}^1;...;\bm{z}_\text{CLS}^k \right] &= e_\text{CLS}, \nonumber \\
    \left[\bm{z}_{\text{i,j}}^1;...;\bm{z}_{\text{i,j}}^k \right] &= \tanh \left( e_{i,j} \right), \nonumber \\
   \mathrm{P}\left(x|e_\text{CLS}, e_\text{i,j}\right) &= \sigma \left( \sum_{m=1}^k \bm{z}_\text{CLS}^{m\intercal} \bm{W}_x^m \bm{z}_{\text{i,j}}^m + b_x \right) \nonumber
\end{align}
where $\mathrm{P}\left(x|e_\text{CLS}, e_\text{i,j}\right)$ represents the probabilities of neutrality, entailment, or contradiction for a span.

\subsection{Logical reasoning over spans}
\label{sec:approach-rules}
Finally, we apply the logical rules for training and inference on fact atoms proposed by \citet{stacey-etal-2024-atomic}. This way, each span’s classification directly informs the final prediction, ensuring the interpretability of the model by tracing every prediction explicitly back to concrete spans.\\

\textbf{Training:} if a given hypothesis is neutral, all extracted spans are to be labelled as neutral.
If it is entailed, the salient spans are to be labelled as entailments, while all other spans are to be labelled neutral.
If the hypothesis is contradicted, the salient spans are to be labelled as contradictions, while any other extracted spans (fact spans) are masked from the loss calculation.
This is because, in the case of a contradiction, other atomic facts might still be in agreement with the hypothesis.\\

\textbf{Inference:} only if the initial global prediction is that the hypothesis is contradicted \textit{and} a contradiction is found among the spans, will the final prediction be contradiction.
Similarly, only if the initial global prediction is that the hypothesis is entailed \textit{and} any span is found to be in agreement, will the final prediction be entailment.
In all other cases, neutral is returned as the prediction.
This ensures that any prediction is faithfully interpretable in the sense that it can be traced back to a specific span in the text.

\subsection{Loss functions and Negative Sampling}

The training loss consists of a total of four separate loss calculations:
For the span extraction, two loss values are calculated for the predictions of $P(\text{is\_left})$ and $P(\text{is\_span})$ using binary cross-entropy loss.
For the initial global classification, as well as the span-wise classification, we apply adaptive thresholding loss \cite{zhou2021atlop}, which is an effective means of managing class imbalances towards a single majority class (neutral applies to most atomic facts) during training, as used in relation extraction (to balance the most common ``no relation'' class).\\

To accelerate training, gold spans are always included as positives, and we additionally sample 50 random spans per instance as negatives for span classification and span-wise NLI. For span-wise NLI, a sampled span is re-labeled as positive if at least $80\%$ of its tokens fall within a salient span.

\section{Experiments}

\subsection{Datasets and Baselines}
Since it is our main baseline of interest, we replicate the experiment setup for FGLR \cite{stacey-etal-2024-atomic} as closely as possible for our evaluation.
This involves using ANLI \cite{nie-etal-2020-adversarial} for training and in-distribution evaluation, as well as out-of-distribution evaluations on ConTRoL \cite{liu2020natural}, RTE \cite{wang-etal-2018-glue}, and WNLI \cite{wang2019glue, levesque2011winograd}. Finally, we add the HANS dataset \cite{mccoy-etal-2019-right} to our evaluation in order to examine the robustness of our models, as this has been a concern with extractive rationale-based models in NLI.\\

We include the following baselines grouped by how fine-grained their interpretability mechanisms are:
For uninterpretable methods we report the results of the encoder-LM fine-tuned on the ANLI dataset\footnote{For consistency, we report the scores reported by \citet{stacey-etal-2024-atomic}, which we were able to reproduce with minor differences attributable to differences in random seeds.}.
For sentence atom interpretability, we include SenLR\footnote{Equivalent to using FGLR without an LLM at inference time and substituting premise sentences for the generated facts.} \cite{stacey-etal-2024-atomic} and a variant of JEDI, JEDI$_\text{sent}$, which uses sentence boundaries provided in the input instead of learning span extraction for span embeddings\footnote{Note that in contrast to SenLR, we provide a supervision signal for salient sentences based on our synthetic rationales.}.
We group together span and fact atoms as the most relevant approaches to compare to our JEDI:
We include SLR-NLI \cite{stacey-etal-2022-logical}, which provides span-level interpretability on the hypothesis using noun phrases as spans, and represents our primary LLM-free, span-level baseline. Naturally, as we aim to distill its behavior into JEDI, we include the results reported by \citet{stacey-etal-2024-atomic} for FGLR, which is the only baseline method requiring an LLM at inference time and for which our method represents the extractive counterpart.
Finally, we evaluate SYRP$_\text{FT}$, the token classifiers described in \ref{sec:syrp-ft}, which allow for token-level interpretability.
SYRP$_\text{FT}$ being extractive but lacking atomic fact decomposition, directly tests our hypothesis that decomposition itself, rather than abstraction over extraction, drives improved robustness.

\begin{table*}[!ht]
    \centering
    \begin{tabular}{lccccccc}
        \hline
        &  \multicolumn{4}{|c}{In-distribution} & \multicolumn{3}{|c}{Out-of-distribution} \\

        Model & R1 & R2 & R3 & ANLI-all & ConTRoL & RTE & WNLI \\
        \hline
        \textit{not interpretable:} & & & & & & & \\
        \debertalarge & 78.3\% & 66.5\% & 61.7\% & 68.1\% & 56.0\%& 90.4\%&  68.9\%\\
        \hline
        \textit{sentence atoms:} & & & & & & & \\
        SenLR & 76.7\% & 64.8\% & 62.0\% & 67.5\% & 56.3\%&  86.3\%&  64.5\%\\
        JEDI$_\text{sent}$ (ours) & \textbf{77.4\%} & \textbf{65.1\%} & \textbf{62.3\%} & \textbf{67.9\%} & \textbf{57.6\%} & \textbf{90.6\%} & \textbf{67.8\%} \\
        \hline
        \textit{span/fact atoms:} & & & & & & & \\
        SLR-NLI & 74.7\% & 60.4\% & 58.3\% & 64.1\% & \textbf{54.7}\%&  87.5\%&  65.8\%\\
        JEDI (ours) & \textbf{75.5\%} & \textbf{63.1\%} & \textbf{59.4\%} & \textbf{65.6\%} & 54.3\% & \textbf{87.7\%} & \textbf{73.7\%}\\
        \hdashline
        FGLR (+GPT-3.5-turbo) & 76.2\% & 64.8\% & 63.1\% & 67.7\% & 52.7\%&  82.0\%&  77.0\%\\
        \hline
        \textit{token atoms:} & & & & & & &\\
        SYRP$_\text{FT}$ (ours) & 75.9\% & 63.3\% & 59.3\% & 65.8\% & 46.3\% &  88.8\% &  65.3\% \\
        \hline

    \end{tabular}
    \caption{Test set scores for \debertalargenospace. Results are averaged accuracies across 10 random seeds.}
    \label{tab:results-large}
\end{table*}

\subsection{Implementation Details}
We train models based on \debertabase and \debertalarge \cite{he2021debertav3} implemented using Huggingface's Transformers \cite{wolf_transformers_2020} and trained using mixed precision. We use AdamW \cite{loshchilov_decoupled_2019} as optimizer (learning rates $\in [7\mathrm{e}{-6}, 9\mathrm{e}{-6}, 1\mathrm{e}{-5}, 3\mathrm{e}{-5}, 5\mathrm{e}{-5}]$ for the encoders, and $1\mathrm{e}{-4}$ for all other parameters). Final hyperparameters were chosen empirically based on validation performance, ensuring a fair comparison across models. We train using linear warmup (1 epoch) \cite{goyal_accurate_2017} followed by a linear learning rate decay. We train each model for 25 epochs and perform early stopping based on development set accuracy.

\subsection{Results and Discussion}

The overall results for \debertalarge are shown in Table \ref{tab:results-large} while those for \debertabase are shown in Table \ref{tab:results-base}.
In Table \ref{tab:HANS}, we further show the results for the HANS dataset, while Table~\ref{tab:ablations} contains results of an ablation study.
Below, we summarize our key findings based on the research questions stated earlier.\\

\noindent
\textbf{JEDI is capable of joint fact decomposition and inference.} While JEDI's accuracy ($65.6\%$) does not quite match that of FGLR ($67.7\%$), which uses an LLM at inference time, it exceeds that of the span-level baseline not relying on generative models ($64.1\%$, SLR-NLI). This demonstrates that atomic decomposition can indeed be effectively distilled into encoder-only architectures, addressing scalability and interpretability without compromising significantly on performance.\\

\noindent
In terms of accuracy, \textbf{sentence atoms are a strong alternative to fact atoms.} Consistent with prior work by \citet{stacey-etal-2024-atomic}, we find sentence-level interpretability to yield high accuracies ($67.9\%$ for JEDI$_\text{sent}$ and $67.5\%$ for SenLR). This finding suggests that if span- or fact-level granularity are not essential, sentence-level supervision provides a highly competitive and simpler alternative.\\

\noindent
\textbf{Synthetic rationales are sufficiently high quality to act as supervision signals.} All our presented approaches rely on the synthetic supervision signals created for SYRP. The overall competitiveness of results when compared to state-of-the-art methods indicates that this does not negatively impact performance. We conclude that the annotations are of sufficiently high quality to act as supervision signals. Importantly, this does not imply that the LLMs used for annotation themselves perform strong extractive NLI: the annotation model was provided with the gold instance label and only tasked with highlighting supporting or contradicting spans.\\

\begin{table}[]
    \centering
    \begin{tabular}{c|cc}
    \hline
    Model    & Acc.$_\text{BASE}$ & Acc.$_\text{LARGE}$ \\
    \hline
       SYRP$_\text{FT}$  & $31.6\%$ & $33.0\%$\\
       JEDI  & $75.0\%$ & $76.9\%$\\
       JEDI$_\text{sent}$  & $80.6\%$ & $83.1\%$\\
    \hline
    \end{tabular}
    \caption{Accuracies measured for models on the HANS dataset, designed to detect whether NLI models rely on shallow syntactic heuristics. The results show clearly that JEDI is more robust than SYRP$_\text{FT}$. For this evaluation we used the models with the highest accuracy on the development split of ANLI.}
    \label{tab:HANS}
\end{table}
\noindent
\textbf{JEDI improves robustness by reducing reliance on shallow heuristics.} Though SYRP$_\text{FT}$, which uses salient token classification to perform NLI, performs on par or even slightly better on in-distribution data ($65.8\%$), JEDI generalizes better to out-of-distribution data: The evaluation on the HANS dataset \cite{mccoy-etal-2019-right}, which is specifically designed to assess whether NLI models rely on shallow syntactic heuristics is presented in Table~\ref{tab:HANS}. The strikingly low scores for SYRP$_\text{FT}$ further emphasize that it is much less robust, which is in line with previous researchers' finding on extractive rationale supervision. The fact that JEDI$_\text{sent}$ scores even higher on HANS suggests that a part of the robustness originates from the span-wise inference architecture, and not just the fact decomposition. This robustness also extends to ConTRoL and WNLI, where JEDI consistently outperforms SYRP$_\text{FT}$, reinforcing the interpretation that its span-wise reasoning architecture at least partially mitigates shortcut learning behaviors.

\subsection{Ablations}

\begin{table}[]
    \centering
    \begin{tabular}{l|ccc}
    \hline
    Model    & interp.? & ANLI & HANS \\
    \hline
        JEDI  & \cmark & $65.6\%$ & $76.9\%$\\
        \hline
        \debertalargecompact & \xmark & $68.1\%\uparrow$ & $79.1\%\uparrow$ \\
        JEDI$_\text{global only}$ & \xmark & $68.2\%\uparrow$ & $80.1\%\uparrow$ \\
        \hline
        JEDI$_\text{no global}$ & \cmark & $59.7\%\downarrow$ & $73.3\%\downarrow$ \\
        JEDI$_\text{w/o ATLoss}$ & \cmark & $65.3\%\downarrow$ & $74.5\%\downarrow$ \\
        \hline
        JEDI$_\text{sent}$ & \cmark$\downarrow$  & $67.9\%\uparrow$ & $83.1\%\uparrow$\\
        SYRP$_\text{FT}$ & \cmark$\uparrow$  & $65.8\%\uparrow$ & $33.0\%\downarrow$\\
    \hline
    \end{tabular}
    \caption{Results of ablation study for models with \debertalarge as backbone. \textit{interp.?} indicates whether interpretability is preserved despite the changes, with \xmark\text{ } indicating no interpretability, and \cmark($\uparrow/\downarrow$) indicating interpretability at a higher or lower level of detail. Arrows $\uparrow \downarrow$ indicate direction of changes over JEDI.}
    \label{tab:ablations}
\end{table}
\begin{table*}[!ht]
    \centering
    \begin{tabular}{lccccccc}
        \hline
        &  \multicolumn{4}{|c}{In-distribution} & \multicolumn{3}{|c}{Out-of-distribution} \\
        Model & R1 & R2 & R3 & ANLI-all & ConTRoL & RTE & WNLI \\
        \hline
        \textit{not interpretable:} & & & \\
        \debertabase & 71.2\% & 54.0\% & 51.7\% & 58.5\% & 53.7\% &  85.0\% &  59.6\% \\
        \hline
        \textit{sentence atoms:} & & & \\
        SenLR & 71.5\% & 55.0\% & \textbf{52.3\%} & 59.1\% & \textbf{53.4\%} &  83.7\% &  53.8\% \\
        JEDI$_\text{sent}$ (ours) & \textbf{72.0\%} & \textbf{55.5\%} & 52.1\% & \textbf{59.4\%} & 52.6\% &  \textbf{85.0\%} &  \textbf{62.4\%} \\
        \hline
        \textit{span/fact atoms:} & & & \\
        SLR-NLI & 65.5\% & 47.8\% & 47.1\% & 53.0\% & 48.9\% &  82.3\% &  56.3\% \\
        JEDI (ours) & \textbf{69.0\%} & \textbf{53.2\%} & \textbf{50.1\%} & \textbf{57.0\%} & \textbf{49.3\%} &  \textbf{83.3\%} &  \textbf{58.5\%} \\
        \hdashline
        FGLR (+GPT-3.5-turbo) & 71.8\% & 56.1\% & 55.3\% & 60.7\% & 49.1\% &  80.8\% &  70.7\% \\
        \hline
        \textit{token atoms:} & & & \\
        SYRP$_\text{FT}$ (ours) & 69.2\% & 53.0\% & 51.7\% & 57.6\% & 49.5\% &  82.2\% &  55.7\% \\
        \hline

    \end{tabular}
    \caption{Test set scores for \debertabasenospace. Results are averaged accuracies across 10 random seeds.}
    \label{tab:results-base}
\end{table*}

In Table~\ref{tab:ablations} we report ablation results. Using only global classification (JEDI$_\text{global only}$) yields performance nearly identical to a standard sequence classifier (\debertalargenospace), showing that the additional span-level losses do not substantially affect global classification. Removing global classification (JEDI$_\text{no global}$), however, causes clear drops, indicating its importance. Replacing ATLoss with cross-entropy (JEDI$_\text{w/o ATLoss}$) leads to modest decreases, suggesting ATLoss is beneficial though not decisive. Finally, JEDI$_\text{sent}$ and SYRP$_\text{FT}$ illustrate the trade-off between accuracy and interpretability granularity: sentence-level aggregation improves accuracy at the cost of coarser explanations, while token-classification substantially harms robustness (accuracy on HANS) despite providing finer interpretability.

\section{Conclusion}
We introduced JEDI, a joint encoder-only architecture capable of performing atomic fact decomposition and interpretable inference in NLI tasks without relying on large generative models at inference time. Our experiments confirmed that JEDI effectively balances interpretability, robustness, and scalability, outperforming span-level baselines and substantially reducing reliance on shallow heuristics. Furthermore, we demonstrated the utility of synthetic rationales, releasing a large-scale corpus (SYRP) to support future interpretability research. Overall, we hope that our work contributes to the development of transparent and scalable NLI systems, highlighting that fine-grained interpretability and robust generalization can be achieved efficiently in encoder-only frameworks.

\section*{Limitations}
JEDI's interpretability relies on extracting and classifying spans from the premise alone, while hypotheses remain undecomposed. Extending JEDI to accommodate atomic decomposition of multi-sentence hypotheses would be necessary for applying the method more broadly. We note that, given appropriate supervision data, our architecture is, in theory, easily expandable to this case, since it is modeled on relation extraction, where classification between spans (atoms) is an inherent part of the task.

Next, our approach relies heavily on synthetic rationales generated by large language models, which may have introduced inaccuracies, despite our targeted model selection procedure designed to minimize such risks. These inaccuracies could potentially propagate errors into the model's interpretability, especially when used on datasets or domains distinct from those evaluated here.

Further, while extractive fact decomposition has clear benefits in terms of computational demands and traceability to explicit text segments, abstractive decomposition can yield paraphrases that are in some cases more natural or easier to interpret. However, abstractive methods are prone to hallucination and may compromise factual accuracy, whereas extractive methods remain grounded in the source text. The relative merits of clarity versus precision are highly context-dependent and not easily captured by current benchmarks, making a fair comparison difficult without carefully designed user studies for specific use cases. Exploring this trade-off in applied settings such as fact-checking constitutes an important avenue for future work and is outside the scope of the present study.

Lastly, our empirical evaluations focus primarily on a limited set of English-language datasets in a small selection of domains. It remains unclear how JEDI would perform on languages or linguistic structures significantly different from those in our evaluation. Specialized domains such as medical NLI were also not evaluated in this work. Addressing multilingual capabilities and broader linguistic and domain coverage constitutes an important direction for future research.

\section*{Acknowledgements}
The authors acknowledge support by the state of Baden-Württemberg through bwHPC.
We thank our reviewers for their helpful feedback and suggestions that helped us refine this manuscript.

\bibliography{anthology,custom}

\appendix
\section{Synthetic Rationale Generation Prompt Design and Model Selection}
\label{sec:appendix-syrp}

\begin{figure*}[]
\centering
\tiny
\begin{tcolorbox}[title=Annotation prompt for entailed hypotheses (used with Qwen/Qwen2.5-32B-Instruct-GPTQ-Int4), fonttitle=\bfseries, colback=gray!10, colframe=black, width=0.95\linewidth, listing only, listing options={
    basicstyle=\ttfamily\small,
    breaklines=true,
    showstringspaces=false
}]
\begin{verbatim}
<|im_start|>system
You are a helpful assistant. You highlight information in text.<|im_end|>
<|im_start|>user
Highlight in the following text any passages supporting the statement that "The trolleybus system has over 2 urban
routes":

The Parma trolleybus system (Italian: "Rete filoviaria di Parma" ) forms part of the public transport network of
the city and "comune" of Parma, in the region of Emilia-Romagna, northern Italy. In operation since 1953, the
system presently comprises four urban routes.

Make only minimal changes, keep everything else unchanged.<|im_end|>
<|im_start|>assistant
```json
{
    "explanation": "The text states that the Parma trolleybus system has *four urban routes*. This supports the
    statement that the trolleybus system has over 2 urban routes.",
    "phrases_to_highlight": ["the system presently comprises four urban routes"],
    "highlighted_text": "The Parma trolleybus system (Italian: \"Rete filoviaria di Parma\" ) forms part of the 
    public transport network of the city and \"comune\" of Parma, in the region of Emilia-Romagna, northern Italy.
    In operation since 1953, *the system presently comprises four urban routes*."
}
```<|im_end|>
<|im_start|>user
Wow! Great job! Let's try another one:

Highlight in the following text any passages supporting the statement that "{hypothesis}":

{premise}

Make only minimal changes, keep everything else unchanged.<|im_end|>
\end{verbatim}
\end{tcolorbox}
\caption{Prompt used for annotating entailed hypotheses SYRP corpus.}
\label{fig:llm-prompt-entailed}
\end{figure*}

\begin{figure*}[]
\centering
\tiny
\begin{tcolorbox}[title=Annotation prompt for contradicted hypotheses (used with Qwen/Qwen2.5-32B-Instruct-GPTQ-Int4), fonttitle=\bfseries, colback=gray!10, colframe=black, width=0.95\linewidth, listing only, listing options={
    basicstyle=\ttfamily\small,
    breaklines=true,
    showstringspaces=false
}]
\begin{verbatim}
<|im_start|>system
You are a helpful assistant. You highlight information in text.<|im_end|>
<|im_start|>user
Highlight in the following text any passages supporting the statement that "Jesse James was a guerrilla in the
Union army during the American Civil War.":

The Centralia Massacre was an incident during the American Civil War in which twenty-four unarmed Union soldiers
were captured and executed at Centralia, Missouri on September 27, 1864 by the pro-Confederate guerrilla leader
William T. Anderson. Future outlaw Jesse James was among the guerrillas.<|im_end|>
<|im_start|>assistant
```json
{
    "explanation": "The text states that Jesse James was among the *pro-Confederate* guerrillas. This
    contradicts the statement that Jesse James was a guerrilla in the Union army.",
    "phrases_to_highlight": ["pro-Confederate guerilla", "Jesse James was among the guerrillas"],
    "highlighted_text": "The Centralia Massacre was an incident during the American Civil War in which twenty-four
    unarmed Union soldiers were captured and executed at Centralia, Missouri on September 27, 1864 by the 
    *pro-Confederate guerrilla* leader William T. Anderson. Future outlaw *Jesse James was among the guerrillas*."
}
```<|im_end|>
<|im_start|>user
Wow! Great job! Let's try another one:

Highlight in the following text any passages contradicting that "{hypothesis}":

{premise}

Make only minimal changes, keep everything else unchanged.<|im_end|>
\end{verbatim}
\end{tcolorbox}
\caption{Prompt used for annotating contradicted hypotheses SYRP corpus.}
\label{fig:llm-prompt-contradicted}
\end{figure*}

Span-based, as well as natural language explanations were generated via prompting instruction-tuned LLMs.
Here, we describe the hyperparameters considered.\\

\noindent
\textbf{Choice of Language Models.}
In order to ensure reproducibility, we opt to use only those large language models for which weights are openly accesible.
Furthermore, we limit the maximum model size to approx. 70 billion parameters due to hardware and compute time constraints.
The above criteria, as well as general benchmark performance of various models result in the following selection of 15 models:
\begin{itemize}
  \item \textbf{Llama3.1} 70B- and 8B-Instruct \cite{grattafiori2024llama3herdmodels}, Tulu-3-70B \cite{lambert2024tulu3}, Nemotron-70B-Instruct \cite{wang2024helpsteer2preferencecomplementingratingspreferences}
  \item \textbf{Llama3.2} 3B- and 1B-Instruct \cite{grattafiori2024llama3herdmodels}
  \item \textbf{Qwen2.5} 72B-, 32B-, 14B-, 7B-, 3B-, 1.5B-, 0.5B-Instruct \cite{qwen2, qwen2.5}
  \item \textbf{Ministral} 8B-Instruct \footnote{\url{https://huggingface.co/mistralai/Ministral-8B-Instruct-2410}}
  \item \textbf{Mistral} Nemo-Instruct \footnote{\url{https://huggingface.co/mistralai/Mistral-Nemo-Instruct-2407}}
\end{itemize}

For larger models, we also perform experiments using quantized variants.\\

\noindent
\textbf{Task Framing.}
In general, we pose the task by providing premise, hypothesis, as well as the label (entailment or contradiction) and request an explanation following a specific format.
Generating natural language explanations is relatively straightforward to elicit from instruction-tuned models by requesting it in a prompt, as it is a generative task.
For span-based explanations, however, the task is not primarily generative, with token classification being a more natural framing. We design two different generative task settings to evaluate: 
(1) \textit{highlighting}, in which the prompt requests for the model to generate the premise text while highlighting the most salient spans, similar to the task given to annotators of the e-SNLI dataset \cite{NIPS2018_8163}.
(2) \textit{redaction}, in which the prompt requests for the model to generate the premise text while redacting any passages entailing or contradicting the hypothesis.\\

\noindent
\textbf{Output Format.}
In order to make the generated outputs easily parseable, the prompts include instructions on how to format the response. Here we examine Markdown and JSON as two possible formats.\\

\noindent
\textbf{Example Placement.}
All experiments are conducted in a one-shot in-context learning setting. Since all the used models allow for three roles, system, user, and assistant, this gives us two fundamentally different options for the placement of the example: (1) The example can either be included in the context as part of the system message, or (2) as a query by a user, followed by a response from the assistant containing the correct solution.\\

\noindent
\textbf{Development Dataset.}
In order to compare different choices of hyperparameters, we manually annotate a dataset consisting of 100 data points (50 of each for cases of entailment and contradiction) taken from the training set of ANLI.\\

\noindent
\textbf{Evaluation of Span-based Rationales.}
We measured the amount of exact matches (which reached a maximum of $30\%$), average intersection-over-union (which reached a maximum of roughly $70\%$) and the frequency with which a given model produces annotations that exceed different threshold values for intersection-over-union.\\

\noindent
\textbf{Evaluation of Natural Language Rationales.}
The prompts were structured to include natural language rationales, not intended as a supervision signal, but as an evaluation tool to check how well a given model interprets the data at hand.
We manually assigned binary labels (acceptable/unacceptable) to $30$ explanations per model.
The top performing combination of model and prompt, which we ended up using, produced a score of $28/30$ acceptable explanations.\\

\noindent
\textbf{Final annotation setup and prompts.}
Figures~\ref{fig:llm-prompt-entailed} and \ref{fig:llm-prompt-contradicted} show the bests prompts chosen for annotation. All other prompts and more details are available at \url{https://jedi.nicpopovic.com}.

\section{SYRP Corpus}
\label{sec:appendix-syrp-stats}

Table \ref{tab:dataset-stats} contains statistics for the entire corpus, while Tables \ref{tab:dataset-stats-train} and \ref{tab:dataset-stats-validation} contain statistics for the train and validation split, respectively.
The dataset is available and an interactive data viewer can be found at \url{https://jedi.nicpopovic.com}.

\begin{table*}[]
\centering
\begin{tabular}{l|c|r|r|r|r|r|r|r}
\hline
Source & Domain & Samples & Entail. & Neutral & Contra. & $\frac{\text{words}}{\text{premise}}$ & $\frac{\text{spans}}{\text{premise}}$ & Coverage \\ \hline
ANLI & Wiki & 162,170 & 51,178 & 69,857 & 41,135 & 54.2 & 1.27 & 19.3\% \\
ConTRoL & Exams &7,114 & 2,618 & 2,183 & 2,313 & 439.2 & 1.53 & 14.5\% \\
ContractNLI & Legal & 7,959 & 3,898 & 3,243 & 818 & 1642.7 & 1.42 & 7.6\% \\
ESNLI & Captions &539,397 & 180,937 & 185,999 & 172,461 & 12.9 & 1.10 & 61.1\% \\
FEVER & Wiki &223,759 & 127,497 & 42,305 & 53,957 & 60.3 & 1.51 & 26.7\% \\
LINGNLI & Diverse &51,167 & 16,940 & 17,438 & 16,789 & 19.6 & 1.08 & 54.0\% \\
MNLI & Diverse &404,827 & 134,907 & 137,152 & 132,768 & 19.9 & 1.12 & 57.1\% \\
WANLI & Diverse &101,180 & 37,099 & 48,977 & 15,104 & 17.5 & 1.03 & 62.7\% \\
\hline
\textbf{Total} & & 1,497,573 & 555,074 & 507,154 & 435,345 & 37.6 & 1.20 & 49.1\% \\
\hline
\end{tabular}
\caption{Overview of the full SYRP-corpus. Average span statistics have been calculated under omission of neutral samples, which do not have any annotated spans. The total number of samples with rationale spans is $991,628$.}
\label{tab:dataset-stats}
\end{table*}

\begin{table*}[]
\centering
\begin{tabular}{l|r|r|r|r|r|r|r}
\hline
Source & Samples & Entail. & Neutral & Contra. & $\frac{\text{words}}{\text{premise}}$ & $\frac{\text{spans}}{\text{premise}}$ & Coverage \\ \hline
ANLI & 159,091 & 50,177 & 68,789 & 40,125 & 54.1 & 1.27 & 19.4\% \\
ConTRoL & 6,344 & 2,344 & 1,946 & 2,054 & 453.9 & 1.53 & 14.4\% \\
ContractNLI & 6,975 & 3,418 & 2,820 & 737 & 1632.6 & 1.41 & 7.4\% \\
ESNLI & 529,905 & 177,680 & 182,764 & 169,461 & 12.9 & 1.10 & 61.1\% \\
FEVER & 204,796 & 121,289 & 35,639 & 47,868 & 60.7 & 1.51 & 26.8\% \\
LINGNLI & 43,918 & 14,446 & 14,995 & 14,477 & 19.5 & 1.08 & 53.9\% \\
MNLI & 385,499 & 128,076 & 130,900 & 126,523 & 19.9 & 1.12 & 56.9\% \\
WANLI & 101,180 & 37,099 & 48,977 & 15,104 & 17.5 & 1.03 & 62.7\% \\
\hline
\textbf{Total} & 1,437,708 & 534,529 & 486,830 & 416,349 & 36.5 & 1.19 & 49.3\% \\
\hline
\end{tabular}
\caption{Overview of the \textbf{training split} of the SYRP-corpus. Average span statistics have been calculated under omission of neutral samples, which do not have any annotated spans. The total number of samples with rationale spans is $951,721$.}
\label{tab:dataset-stats-train}
\end{table*}

\begin{table*}[]
\centering
\begin{tabular}{l|r|r|r|r|r|r|r}
\hline
Source & Samples & Entail. & Neutral & Contra. & $\frac{\text{words}}{\text{premise}}$ & $\frac{\text{spans}}{\text{premise}}$ & Coverage \\ \hline
ANLI & 3,079 & 1,001 & 1,068 & 1,010 & 54.5 & 1.25 & 17.3\% \\
ConTRoL & 770 & 274 & 237 & 259 & 317.9 & 1.55 & 14.9\% \\
ContractNLI & 984 & 480 & 423 & 81 & 1714.6 & 1.47 & 9.5\% \\
ESNLI & 9,492 & 3,257 & 3,235 & 3,000 & 13.9 & 1.12 & 59.6\% \\
FEVER & 18,963 & 6,208 & 6,666 & 6,089 & 55.7 & 1.39 & 24.7\% \\
LINGNLI & 7,249 & 2,494 & 2,443 & 2,312 & 19.8 & 1.08 & 54.6\% \\
MNLI & 19,328 & 6,831 & 6,252 & 6,245 & 19.5 & 1.10 & 59.7\% \\
\hline
\textbf{Total} & 59,865 & 20,545 & 20,324 & 18,996 & 63.6 & 1.21 & 44.7\% \\
\hline

\end{tabular}
\caption{Overview of the \textbf{validation split} of the SYRP-corpus. Average span statistics have been calculated under omission of neutral samples, which do not have any annotated spans. The total number of samples with rationale spans is $39,907$.}
\label{tab:dataset-stats-validation}
\end{table*}

\end{document}